# A Swift Heuristic Method for Work Order Scheduling under the Skilled-Workforce Constraint

Nima Safaei[a], Corey Kiassat[b]

[a] *Data Science and Analytics Lab, Global Banking and Marketing, Scotia Bank, Toronto, ON, Canada*

*nima.safaei@scotiabank.com*

[b] *Assistant Professor of Industrial Engineering, Department of Engineering, Quinnipiac University, 275 Mt. Carmel Avenue, Hamden, CT 06518, USA*

*corey.kiassat@qu.edu*

**ABSTRACT:** The considered problem is how to optimally allocate a set of jobs to technicians of different skills such that the number of technicians of each skill does not exceed the number of persons with that skill designation. The key motivation is the quick sensitivity analysis in terms of the workforce size which is quite necessary in many industries in the presence of unexpected work orders. A time-indexed mathematical model is proposed to minimize the total weighted completion time of the jobs. The proposed model is decomposed into a number of single-skill sub-problems so that each one is a combination of a series of nested binary Knapsack problems. A heuristic procedure is proposed to solve the problem. Our experimental results, based on a real-world case study, reveal that the proposed method quickly produces a schedule statistically close to the optimal one while the classical optimal procedure is very time-consuming.

**INTRODUCTION**

Organizations frequently aim to optimally manage their available resources to minimize operations costs and maximize resource utilization and asset availability. The authors' experiences from several research projects in the operations scheduling field show that maintenance scheduling, given limited resources over a short planning horizon, is an important challenge at the operational level; this agrees with a review paper by Mendez *et al.* (2006). In these research projects, one of the most important maintenance resources is *skilled workforce*. Maintenance jobs are labour intensive, and the workforce performing these jobs is highly-paid as a result of being extremely skilled in the respective areas.

The execution of the scheduling process has been a tedious and time-consuming job, and is often a highly intricate job when it deals with employees who have specific skills, job grade, and working on different shift patterns. The Scheduling Problem under the Skilled-workforce

Constraint (SPSC) has wide applications in various industries to manage maintenance jobs and workload. Scheduling of maintenance jobs associated with steel production machinery (Safaei *et al.,* 2011a), military aircraft fleet (Safaei *et al.,* 2011b), commercial aircraft (De Bruecker *et al.,* 2015b) or power transmission equipment (Safaei *et al.* 2012b) are prominent examples where limited labour resource and skilled-workforce availability is of great importance. Moreover, multiple skills have gained a lot of attention in recent studies related to other fields (Heimerl and Kolisch, 2010; De Bruecker *et al.,* 2015a; De Bruecker *et al.,* 2015b).

One difficulty with SPSC is how to handle the uncertainty due to unpredictable events such as unexpected equipment shutdown, power interruption or aircraft breakdown. These situations do not commonly occur with a known pattern. Therefore, such uncertainty leads maintenance schedulers to seek a list of alternative schedules and various what-if scenarios under different workforce sizes. This is especially true when a part of labour requirements may be satisfied through external resources. A further complication is the frequent usage of a built-in buffer of labour for catastrophic failures or emergency situations. Therefore, there is a serious need in some industries for a quick and efficient method to solve an SPSC and especially for a sensitivity analysis on the skilled workforce availability (Keysan *et al.,* 2010). For example, in military jets where we encounter daily missions and high frequency of unexpected faults, especially during wartime, sufficient labour resource should be available simultaneously in both flight line and repair shop (Safaei *et al.* 2012b). Thus, the maintenance scheduler needs to know the possible what-if scenarios to exchange the labour between flight line and shop in the presence of unexpected failures. Likewise, the maintenance department in a steel production has to re-prioritize the maintenance jobs, arriving from different manufacturing areas, and re-distribute the technicians among different areas when a critical failure is reported (Safaei *et al.,* 2011a). This research work is motivated by several research projects which have a strong reliance on having an efficient tool to quickly solve an SPSC with the aim of generating what-if scenarios and performing the sensitivity analysis on the skilled-workforce availability. The reason for this reliance is frequent stochastic events in the form of unscheduled repair jobs.

SPSC aims to match a series of jobs with a set of skills. Each job has certain requirements that are to be completed by various skills. The requirements consist of the processing time and the number of required technicians, assumed to be known in advance. The available number of technicians of each skill is also known and will remain fixed over the planning horizon. As pointed out earlier, the main constraint is the workforce availability per unit time. That is, the number of technicians of each skill required per unit time cannot exceed the number of persons available for that skill, whilst all jobs must be completed by the end of the considered horizon. The jobs are assumed to have different weight/priority and the objective is to minimize the Total Weighted Completion Time (TWCT) of all jobs. The problem is essentially NP-hard and, based on the previous experimental results, the computational efforts will progressively increase when the available workforce size decreases, while the set of tasks remains the same (Safaei et al., 2011a).

A comprehensive literature review on SPSC with a focus on the maintenance field can be found in (Safaei et al., 2011a) and (Safaei et al., 2011b). In the former study, the authors consider a bi-objective SPSC, associated with a steel production company, and formulate it using the flow network with integral constraints. This is done to trace the workforce flow through the jobs. As such, Safaei et al. (2012a) propose a simulated annealing algorithm, with parallel architecture, to solve their previous formulation (Safaei et al., 2012a). The aim is to generate the Pareto set of the alternative solutions. Their approach is able to produce near-optimal Pareto solutions in a reasonable time compared to the optimal Pareto set. However, this approach is not quick enough to be used as a robust post-optimization tool in the presence of unexpected events.

**CONTRIBUTION**

In this paper, a heuristic approach is proposed to solve an SPSC within a very short period of time while the quality of solutions is significantly close to the optimal solution obtained by the classical methods. To this end, we have a new perspective on SPSC in which the planning horizon is divided into a number of nested subintervals so that each subinterval is thought of as a two-dimensional knapsack. The subintervals are arranged in such a way that the inner-most knapsack represents the first time unit and the outer-most one represents the entire horizon. The capacity of each knapsack is equivalent to the available man-hours during the specific subinterval. As such, the jobs are interpreted as bi-dimensional items comprising of labour requirement (human) and processing time (hour) as the two dimensions.

In our strategy, the minimization of TWCT imposes that items with higher priority be included in the inner-most knapsacks. Note that the above strategy differs from the classical decomposition of Lagrangian relaxation problem (Fisher, 1981) in which each time period is considered as a separate 0-1 knapsack problem. Instead, the '*nested*' concept is used to consider both time and labour dimensions at the same time. Consequently, SPSC can be considered as a crossover between the *multi-dimensional* and the *multi-period* knapsack problems, which makes the proposed strategy quite a novel idea. This is due to the consideration of multiple skills as well as having each subinterval as a single period. SPSC might be converted to a specific kind of the Knapsack Problem (KP), such as multi-period KP (Faaland, 1981), multi-dimensional KP (Weingartner and Ness, 1967; Hill and Reilly, 2000); *T*-Constraint KP (Shih, 1979; Pirkul, 1987), or multiple KP with assignment restrictions (Dawande et al., 2000). In doing so, the applied solution approaches cannot be directly used for the SPSC considered in this study because of its particular structure that will be discussed in detail in the next section. SPSC can also be considered as a specific case of Resource-Constrained Project Scheduling Problem (RCPSP) with 'Labour' as a constant and renewable resource (Hartmann, 2000). However, the concept of 'skill' as well as 'TWCT' as an objective makes it impossible to apply the solution methods of RCPSP to SPSC. Most methods have been developed to solve RCPSP considering the minimal Makespan (project duration) as the objective function. Moreover, some classical methods, such as Metra potential or critical path method, do not explicitly take into account the resource constraints (Sprecher *et al.*, 1997). Others have been focused on specific characteristics of RCPSP such as multiple modes, non-renewable resources, time-dependent resources and activity time-windows (Brucker *et al.*, 1999; Brucker and Kunst, 2008). Therefore, it can be concluded that, due to the specific characteristics of the SPSC considered in this study, the approaches introduced in the literature to solve different kinds of KPs and different versions of RCPSP either cannot be customized to solve an SPSC or are not sufficiently quick and efficient to be used as a post-optimization tool.

The key idea behind the proposed heuristic method consists of two phases: at first, the SPSC is decomposed into a number of single-skill sub-problems such that each sub-problem is a combination of a series of nested binary KPs. As a next step, a procedure, inspired by the Dantzig method (Dantzig, 1957), is developed to solve the sub-problems separately. The decision to use the Dantzig method is due to the capability of representing the capacity constraints in KPs as a nested form. Using some propositions, we show how Danzig's strategy can be simply applied to solve the SPSC. The findings reveal that the combination of the partial solutions associated with the sub-problems results in a high quality solution to the SPSC in a very short period of time.

The remainder of the paper is organized as follows. In Section 2, a time-indexed mathematical formulation for the SPSC is introduced. The knapsack formulation, as well as the extension of the Dantzig method, is discussed in Section 3. The proposed solution approach is described in Section 4. Finally, Section 5 verifies the performance of the proposed approach using a set of real data and also covers the related analysis.

**MATHEMATICAL FORMULATION**

In this section, SPSC is formulated as a time-indexed 0-1 mathematical programming model to minimize TWCT. The following assumptions are adopted from a real life case study associated with a steel company in Ontario, Canada (Safaei *et al.*, 2011a). The company has a plant-wide scheduling approach through a central department to respond to maintenance work orders/jobs of various Manufacturing Areas (MAs) in the plant. The aim of this department is to minimize workforce costs and to avoid long-term disruptions and shutdowns of the equipment within MAs. The maintenance jobs are prioritized based on equipment criticality, order due date, and the severity of failure, resulting in a normalized weight assigned to each job (Safaei *et al.*, 2011a).

*Assumptions*

1. Planning Horizon is assumed to be a countable finite set of *T* time units [*t*-1, *t*) where *t* = 1, 2, ..., *T*. The jobs can only start at the beginning of time units, i.e., instances of time *s* = 0, 1,..., *T*-1. In our case, a weekly planning horizon consisting of five working days is considered. Each day has two consecutive 8-hour work shifts with an hour-long break between them.
2. All submitted maintenance jobs are first prioritized to be scheduled over the upcoming day. The goal is to complete all jobs by the end of the day. Jobs which cannot be scheduled over the upcoming day are postponed to the following day as high priority jobs.
3. Due to the high setup/preparation times, job pre-emption is not allowed.
4. Setup time is built into job duration
5. The number of available technicians of each skill designation is determined in advance and remains fixed over the upcoming day. Each technician may have different skills; however, he/she is assigned to work within only one skill designation over the upcoming day.
6. There is no precedence relationship among the skills to perform the jobs.

*Input parameters*

$T$      length of planning horizon in terms of a known time unit *t*, where *t* = 0,1,2,...,*T*

$M$      number of jobs, where *m* = 1,2,...,*M*

$K$      number of skills, where *k* = 1, 2,...,*K*

$p_{mk}$      processing time required for skill *k* to perform job *m*, where $0 \leq p_{mk} \leq T$

$\lambda_{mk}$      required number of technicians of skill *k* to perform job *m*

$b_k$      available number of technicians of skill *k* over the planning horizon

$w_m$      weight or priority of job *m*, where $0 < w_m \leq 1$ and $\sum_{m=1}^{M} w_m = 1$

*Decision Variables*

$y_{mkt}$ =1 if processing of job *m* is started at time instant *t* (or equivalently at the beginning of time unit *t*+1) by skill *k*; and equals zero otherwise.

$\varphi_m$      overall completion time of job *m* by all associated skills, i.e., $\varphi_m = \max_{1 \leq k \leq K} \{c_{mk}\}$, where $c_{mk} = p_{mk} + \sum_{t=0}^{T-1} t y_{mkt}$ represents the completion time of job *m* by skill *k*.

*Mathematical Model*

A Mixed-Integer Programming (MIP) formulation for the SPSC is:

P:

$$\min Z = \sum_{m=1}^{M} w_m \varphi_m \qquad (1)$$

s.t:

$$\varphi_m \geq p_{mk} + \sum_{t=0}^{T-1} t y_{mkt} \qquad \forall m,k; p_{mk} \neq 0 \qquad (2)$$

$$\sum_{t=0}^{T-1} y_{mkt} = 1 \qquad \forall m,k; p_{mk} \neq 0 \qquad (3)$$

$$\sum_{m=1}^{M} \left( \sum_{s=\max\{0, t-p_{mk}\}}^{t-1} y_{mks} \right) \lambda_{mk} \leq b_k \qquad \forall k, 1 \leq t \leq T \qquad (4)$$

$$y_{mkt} \in \{0,1\} \qquad \forall m,k,t$$

The objective function presented in (1) is to minimize TWCT of the jobs so that the overall completion time of each job by all associated skills is determined by (2). TWCT is used to minimize the downtime of equipment due to maintenance or repair jobs (Safaei *et al.*, 2011a). The term $\sum_{t=0}^{T-1} t y_{mkt}$ in (2) represents the start time of job *m* by skill *k*. Assignment constraint (3) imposes the processing of job *m* by skill *k* must be started during the planning horizon. Inequality (4), inspired by the classical resource constraint in RCPSP, guarantees that the required number of technicians of skill *k* in each time-unit cannot exceed the total available number of technicians of that skill, i.e., $b_k$. The term

$$\gamma_{mk}^t = \left( \sum_{s=\max\{0,t-p_{mk}\}}^{t-1} y_{mks} \right)$$

in (4) shows whether job *m* is being processed by skill *k* at time-unit *t*. That is, $\gamma_{mk}^t = 1$ if job *m* is started within interval [max{0, *t*-$p_{mk}$}, *t*). Assumption 2 imposes the following *necessary feasibility condition* for (P)

$$W_k = \sum_{m=1}^{M} p_{mk} \lambda_{mk} \leq T \times b_k, \qquad (5)$$

which means the total man-hours of skill *k* required to perform all jobs, $W_k$, cannot exceed the available man-hours over the horizon, $T \times b_k$. Without loss of generality, we assume the input datasets satisfy the condition above.

From a practical point of view, Model P cannot be solved optimally in a reasonable amount of time for real-sized instances. First of all, P is NP-hard since it is an MIP model (Garey and Johnson, 1979). Moreover, P is an equality-constrained MIP (Eq. 3) and therefore P can also be categorized as an NP-complete problem due to the non-negativity requirement on $y_{mkt}$ (Aardal *et al.*, 2000; Aardal and Lenstra, 2002). It is worth mentioning that P is essentially a nonlinear integer programming problem when constraint (2) is removed and $\varphi_m = \max_{1 \leq k \leq K}\{c_{mk}\}$ directly embedded in (1) as $Z = \sum_{m=1}^{M} w_m \max_{1 \leq k \leq K}\{c_{mk}\}$. Consequently, the MIP formulation may be an alternative to slightly reduce the problem complexity and to guarantee the optimal solution; however, computational time still remains a challenging issue for real size instances. One common alternative to achieve a lower bound on (1) is to decompose P into $K$ subproblems in terms of different skills. By applying the primary inequality $\sum_i \max_j \{a_{ij}\} \geq \max_j \{\sum_i a_{ij}\}; a_{ij} \geq 0$ on the original non-linear objective function, we get the following:

$$Z = \sum_{m=1}^{M} w_m \max_{1 \leq k \leq K} \{c_{mk}\} = \sum_{m=1}^{M} \max_{1 \leq k \leq K} \{w_m c_{mk}\} \geq \max_{1 \leq k \leq K} \left\{ \sum_{m=1}^{M} w_m c_{mk} \right\} = \max_{1 \leq k \leq K} \{Z_k\}, \quad (6)$$

where $Z_k$ represents the weighted completion time of jobs by skill $k$ so that $\max_k \{Z_k\}$ is an explicit lower bound on $Z$. To calculate the components $Z_k$, P can be decomposed into $K$ single-skill sub-problems by relaxing (2), as follows:

P$_k$:

$$\min Z_k = \sum_{m=1}^{M} w_n (p_{mk} + \sum_{t=0}^{T-1} t y_{mkt}), \quad (7)$$

s.t:

$$\sum_{t=0}^{T-1} y_{mkt} = 1 \qquad \forall m; p_{mk} \neq 0, \quad (8)$$

$$\sum_{m=1}^{M} \left( \sum_{s=\max\{0, t-p_{mk}\}}^{t-1} y_{mks} \right) \lambda_{mk} \leq b_k \qquad 1 \leq t \leq T \quad (9)$$

$$y_{mkt} \in \{0,1\} \qquad \forall m, t.$$

(P$_k$) is in fact a Generalized Assignment Problem (GAP) in which $i_k = \sum_{m=1}^{M} \text{sgn}(p_{mk}) \leq M$ items must be assigned to $T$ bins with respect to the capacity constraint (9). The sign function 'sgn' equals 1 if $p_{mk}>0$ and equals 0 otherwise. By dualizing constraint (8), (P$_k$) reduces to $T$ 0-1 classical knapsack problem and the resulting Lagrangian relaxation problem can thus be solved in time proportional to $T \times i_k \times b_k$ (Fisher, 1981). Hence, a lower bound on (1) can be obtained using an exact approach in time proportional to $T \sum_k i_k b_k$. Let $\hat{x}_k$ be the solution of (P$_k$) obtained by any approach so that $\hat{X} = (\hat{x}_1, \hat{x}_2, ..., \hat{x}_K)$ is a feasible solution for (P). According to (6), we have the following:

$$Z(\hat{X}) \geq Z(X^*) \geq \max_{1 \leq k \leq K} \{Z_k(\hat{x}_k)\},$$

where $X^*$ is the global optimal solution for primary model P. As a specific case, if $\hat{x}_k$ is the optimal solution of P$_k$, we will get $X^* = \hat{X}$, when each job needs just one skill, i.e., for each $m$ there is only one $k$, where $p_{mk} > 0$.

The heuristic method proposed in this paper is aimed at obtaining the partial solutions $\hat{x}_k; k=1,2,...,K$, so that the lower and upper bounds in the above inequality are tight enough where $Z(\hat{X})$ will be a good approximation on $Z(X^*)$. This benefit is due to "*nested knapsacks*", the key idea behind the proposed method. Since the Dantzig method is a fast and near-optimal method for the classical KPs, the proposed nested strategy ensures the generated solution has the least volatility to the problem characteristics such as size, variability of processing times, and skill requirements. Thus, the lower and upper bounds in the above inequality are not affected by the problem characteristics, as statistically verified through the computational experiences. Our findings show that the computation effort of the proposed heuristic method is much less than the time proportion $T\sum_k i_k b_k$ while the optimality gap is not statistically significant. Thus, the proposed method can be used as a post-optimization tool to generate the list of what-if scenarios with respect to the different levels of the workforce availability in a very short period of time.

**NESTED KNAPSACK FORMULATION FOR ($P_k$)**

By extracting the constant part of Eq. (7), $Q$, we have:

$$Z_k = \sum_{m=1}^{M} w_m \left( p_{mk} + \sum_{t=0}^{T-1} t y_{mkt} \right) = \sum_{m=1}^{M} w_m p_{mk} + \sum_{m=1}^{M} \sum_{t=0}^{T-1} (w_m t) y_{mkt} = Q + \sum_{t=0}^{T-1} \sum_{m=1}^{M} (w_m t) y_{mkt}.$$

Hereafter, (7) is replaced by the reduced form $\sum_{t=0}^{T-1} \sum_{m=1}^{M} (w_m t) y_{mkt}$. In this reduced form, the coefficient ($w_m t$) is interpreted as the processing cost of job $m$, if it is started at time $t$. This coefficient dictates that jobs with higher priorities should be scheduled in earlier time units to minimize $Z_k$. Recalling the necessary feasibility condition (5), we can express (9) in another way. That is, inequality (5) must hold for each subinterval [0, $t$), where $1 \leq t \leq T$, as follows:

$$\sum_{s=0}^{t-1} \sum_{m=1}^{M} \min\{p_{mk}, t-s\} \lambda_{mk} y_{mks} \leq t \times b_k \quad \forall t \tag{10}$$

This inequality, representing the nested form of capacity constraint (9), states that the total man-hours required to handle the workload allocated to subinterval [0, $t$) cannot exceed the available labour resource $t \times b_k$, in which $t$ is the length of the subinterval. According to the left side of (10), if job $m$ is started somewhere within [0, $t$), i.e., $\sum_{s=0}^{t-1} y_{mks} = 1$, a part of its process with length min{$p_{mk}$, $t$-$s$+1} lies within [0, $t$). Inequality (10) is a special case of (5) in the extreme case of $t=T$, considering (8), and knowing that the allowable interval to start job $m$ is [0, $T$-$p_{mk}$]. Similar to (9), inequality (10) also considers job processing continuity and guarantees the required man-hours not exceeding the available resources per unit time. Hence, (10) can replace (9) in $P_k$.

Inequality (10) in fact represents a series of capacity constraints for $T$ Binary KPs (BKPs) associated with the nested subinterval [0, $t$); $1 \leq t \leq T$. Subinterval [0, $t$) represents the $t$[th] knapsack with capacity $C_t = t \times b_k$ (man-hour). In this case, job $m$ occupies $a_{ms}$ = min{$p_{mk}$, $t$-$s$}×$\lambda_{mk}$ units of capacity of knapsack $t$, if it is started at time instant $s \in [0, t)$. The objective of the $t$[th] KP is to place the most valuable workload in knapsack *t*, or equivalently to schedule the most important jobs within subinterval [0, *t*). The BKP associated with subinterval [0, *t*) is as follows:

BKP$_k$(*t*):

$$\max v(t) = \sum_{s=0}^{t-1} \sum_{m=1}^{M} w_m y_{mks}, \qquad (11)$$

s.t:

$$\sum_{s=0}^{t-1} \sum_{m=1}^{M} a_{ms} y_{mks} \leq t \times b_k.$$

$$y_{mks} \in \{0,1\} \qquad \forall m,s; \ s=0,...,t-1.$$

Using the following propositions, we show that subproblems BKP$_k$(*t*) can be simply solved through the Dantzig method.

**Proposition 1**: when *t* approaches *T*, $a_{ms}$ becomes independent of index *s*, that is

$$t \to T \Rightarrow a_{ms} \to p_{mk} \times \lambda_{mk}.$$

**Proof:** In general, the processing times of the jobs is supposed to be significantly smaller than the length of the planning horizon, i.e, $p_{mk} \leq T$ $\forall m,k$. Hence, as *t* approaches *T*, $a_{ms}$ gradually becomes independent of index *s* and can be estimated by $p_{mk} \times \lambda_{mk}$ □

**Proposition 2**: According to Proposition 1, when *t* approaches *T*, BKP$_k$(*t*) can be rewritten independently of index *s*, as follows:

$$\text{BKP}_k(t): \left\{ \max \sum_{m=1}^{M} w_m x_{mkt} \mid s.t: \sum_{m=1}^{M} (p_{mk} \lambda_{mk}) x_{mkt} \leq t \times b_k; \ x_{mkt} \in \{0,1\} \right\} \qquad (12)$$

where $x_{mkt} = \sum_{s=0}^{t-1} y_{mks} = 1$ if the processing of job *m* is started somewhere within [0, *t*). It equals 0 otherwise.

**Proposition 3**: Dantzig method states that the *optimal fractional solution* to LP relaxation of Problem (12), LP-BKP$_k$(*t*), where the indices of $x_{mkt}$ ($0 \leq x_{mkt} \leq 1$; $pt_{mk} > 0$ $\forall m$) are arranged in non-increasing order of the efficacy ratios $w_m / (p_{mk} \lambda_{mk})$, is given by

$$\begin{cases} x_{mkt} = 1, \text{ if } m < m_0^t \\ x_{mkt} = 0, \text{ if } m > m_0^t \\ x_{mkt} = \left( C_t - \sum_{n < m_0^t} p_{nk} \lambda_{nk} \right) \Big/ (p_{mk} \lambda_{mk}), \text{ if } m = m_0^t, \end{cases} \qquad (13)$$

where $m_0^t$ is the smallest integer $1 \leq m_0^t \leq M$ for which $\sum_{m \leq m_0^t} p_{mk} \lambda_{mk} \geq C_t$, where $C_t = t \times b_k$. If no $m_0^t$ exists, then $x_{mkt} = 1$ $\forall m$. This indicates there is sufficient workforce capacity (of skill *k*) so that the whole workload can be included in knapsack *t*. If $x_{m_0^t kt} = 0$, then the resulting *integer* solution is optimal for LP-BKP$_k(t)$ and BKP$_k(t)$, regardless of the non-assigned jobs $\{m_0^t, m_0^t + 1, ..., M\}$. Proposition 3 reflects the idea of *nested knapsacks*, the key support behind the proposed solution procedure.

**SOLUTION PROCEDURE**

In this section, a fast heuristic method is proposed through Proposition (3) so that at each iteration *t*, Problem (12) is solved using (13) for the jobs which have not been assigned in inner knapsack *t* – 1. Note that Proposition 3 is not always in line with objective TWCT. To minimize TWCT, by relaxing the resource constraints, i.e., $b_k \to \infty$, the jobs should be arranged in non-increasing order of the efficacy ratios $w_m/p_{mk}$ so that $w_{[1]}/p_{[1]k} \geq w_{[2]}/p_{[2]k} \geq \cdots \geq w_{[M]}/p_{[M]k}$. In this case, Proposition 3 will support TWCT if and only if $\lambda_{[1]k} \leq \lambda_{[2]k} \leq \cdots \leq \lambda_{[M]k}$. Proposition 3 simply states that, under the scenario $w_{[1]}/p_{[1]k} \approx w_{[2]}/p_{[2]k} \approx \cdots$, more resource-consuming jobs should be scheduled later. This is intuitive as high resource-consuming jobs act as a bottleneck. To illustrate the issue, consider two typical jobs to be done by a given skill *k* where $p_{1k} = 1, \lambda_{1k} = 5, w_1 = 1; p_{2k} = 2, \lambda_{2k} = 1, w_2 = 1$ and $b_k = 5$. Considering Proposition 3, the optimal sequence is $2 \succ 1$ with $TWFT = 1 \times 2 + 1 \times 3 = 5$. However, using the efficacy ratios $w_m/p_{mk}$, the optimal sequence is $1 \succ 2$ with $TWFT = 1 \times 1 + 1 \times 3 = 4$. Therefore, proposition 3 produces a worse TWCT. By inserting a new job $p_{3k} = 2, \lambda_{3k} = 2, w_3 = 1$, proposition 3 produces the optimal sequence $1 \succ 3 \succ 2$ with a much better TWCT rather than efficacy ratios $w_m/p_{mk}$ because jobs 1 and 3 can be scheduled simultaneously. Even though this strategy may not be in line with TWCT; as our computational analysis reveals, the gap between the optimal solution and our procedure remains relatively constant even if the problem size increases and/or the workforce resource are tightened. This is the main advantage of our procedure and is statistically inferred in Section 5.

To develop our procedure, we first revise (13) to cope with LP constraint $x_{mkt} \in \{0,1\}$. The following is the result:

$$\begin{cases} x_{mkt} = 1, \text{ if } m \leq m_0^t \\ x_{mkt} = 0, \text{ if } m > m_0^t \end{cases}, \quad (14)$$

where $m_0^t$ is the largest integer $1 \leq m_0^t \leq M$ for which $\sum_{m \leq m_0^t} p_{mk} \lambda_{mk} \leq C_t$. Recall the reduced form of (7), coefficient ($w_m t$) imposes that items with higher priorities should be scheduled in the earlier time units, resulting in the inner knapsacks having a higher priority to be packed. However, these inner knapsacks are practically interrelated due to (8), the nested nature of the subintervals, and also the non-pre-emptive assumption. More precisely, each item *m* started at the beginning of time-unit *s* represents a workload equivalent to $p_{mk} \times \lambda_{mk}$ man-hours. In this case, the proportions $\{(r-s)/p_{mk}\}$ of the workload are included in the consecutively nested knapsacks *r*= *s*, *s*+1,…, and *s*+$p_{mk}$-1 respectively. This is demonstrated in Figure 1 where job [3] is included in knapsacks 2, 3, and 4. Thus, a feasible strategy to solve (P$_k$) is to sequentially solve BKP$_k$(*t*); 1≤*t*≤*T*, starting from the first time-unit. That is, at each iteration, the most valuable workload is placed in knapsack *t*, counting the workload already placed within inner knapsacks 1 to *t*-1. Note this strategy dictates that if $m \in \psi_t$ and $x_{mkt} = 1$, we have $y_{mk(t-1)} = 1$. This is because of the reasoning behind coefficient ($w_m t$) which enforces jobs to start at the earliest possible time. In Figure 1, Job [3] is scheduled at the earliest possible time *t*=2, since it cannot be included in the first knapsack. Hence, without loss of generality, variable '*y*' is substituted by '*x*' in (12) and accordingly in (13) and (14) hereafter.

We have developed an extension of the Dantzig Method to implement the aforementioned strategy. We have called this procedure the Extended Dantzig Method, or EDM for short. At each iteration of EDM, BKP$_k$(*t*) is solved using (14). The feasible job indices at each iteration are those which have not been started yet. This is shown by the set $\psi_t = \{m \mid \sum_{s=0}^{t-1} y_{mks} = 0\}$. Moreover, the capacity of the *t*$^{th}$ knapsack is updated as below with respect to the workload already assigned to the inner knapsacks 1 to *t*-1.

$$C_t = t \times b_k - \sum_{s=0}^{t-1} \sum_{m=1}^{M} \min\{p_{mk}, t-s\} \lambda_{mk} y_{mks} \quad \forall t.$$

The main body of the procedure is as follows:

PROCEDURE **EDM**:

Initialize $C_t = 0$, $\psi_t = \varnothing$ $\forall t$;

FOR *t* = 1 TO *T*

a. $C_t = t \times b_k - \sum_{s=0}^{t-1} \sum_{m=1}^{M} \min\{p_{mk}, t-s\} \lambda_{mk} y_{mks}$

b. Set $\psi_t = \{m \mid \sum_{s=0}^{t-1} y_{mks} = 0\}$ and then sort the elements of $\psi_t$ in order of non-increasing ratios $w_m / (p_{mk} \lambda_{mk})$. If $\Psi_t = \varnothing$ Then stop, Else go to step (c);

c. According to (14), set $y_{mkt}$ values;

NEXT *t*

END PROCEDURE

In Figure 1, five jobs numbered in a non-increasing order of ratios $w_m / (p_{mk} \lambda_{mk})$ have been scheduled sequentially using EDM procedure. It is assumed that $b_k$ = 6 people are available over the planning horizon. The number within the boxes represents $\lambda_{mk}$ and the length of each box represents $p_{mk}$. The grey dashed rectangles indicate the nested knapsacks $t$ = 1, 2, 3, and 4 where $t^{th}$ knapsack has a capacity equal to $C_t = t \times 6$.

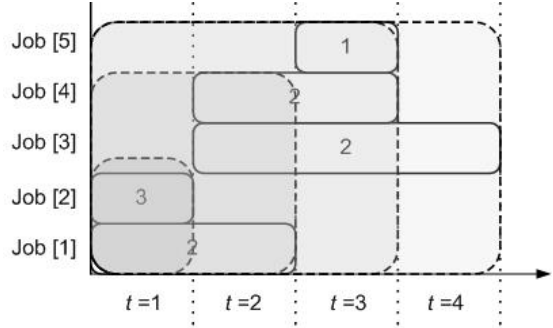

**Figure 1.** Schematic scheduling of five jobs using EDM

There is still a possibility to improve EDM by better utilizing the *remaining capacity* of the knapsacks due to the integrity property of $y_{mks}$ in (14). In other words, at each iteration of EDM, the remaining capacity of the knapsack may still be used by a subset of jobs not selected by (14) due to low efficacy. For example, in Figure 1, even though job (5) has the lowest efficacy, it can be selected at the first iteration to be placed in the first knapsack for improving (11). This will also lead to an improvement of (7). To consider this improvement in the EDM procedure, we conduct a two-level strategy in which, at the first level, the jobs are selected in terms of their efficacy through (14). At the second level, some non-selected items might be picked with the aim of maximizing the capacity utilization and accordingly improving (7). To consider this improvement, (14) is changed to the following:

$$\begin{cases} y_{mkt} = 1, & \text{if } \left[ \left( m \le m_0^t \right) \vee \left( m \in \upsilon_t \right) \right] \\ y_{mkt} = 0, & \text{if } \left[ \left( m > m_0^t \right) \vee \left( m \notin \upsilon_t \right) \right] \end{cases}, \quad (15)$$

where $\upsilon_t$ represents an optimal subset of $R_t = \{m_0^t + 1, ..., M\}$ such that $\sum_{m \in \upsilon_t} p_{mk}\lambda_{mk} \le \theta_t$. $R_t$ represents the set of non-selected jobs and $\theta_t$ represents the remaining capacity of knapsack $t$ after applying (14), where $\theta_t = C_t - \sum_{m \le m_0^t} p_{mk}\lambda_{mk}$. Assuming $\theta_t \ge \min_{1 \le m \le M} \{p_{mk}\}$, there may be the possibility to fill the remaining capacity using the jobs belonging to $\upsilon_t$. $\upsilon_t$ is obtained by solving the following 0-1 KP with the objective of maximizing the capacity utilization.

$$\left\{ \max\left( \sum_{m \in R_t} u_{mk} \right) \text{ s.t: } \sum_{m \in R_t} p_{mk}\lambda_{mk}u_{mk} \le \theta_t; \ u_{mk} \in \{0,1\} \right\}$$

$u_{mk}$ =1 means $m^{th}$ element of $R_t$ is assigned to the remaining capacity and 0 otherwise. Yet again applying (14) to solve the above sub-problem, the jobs belonging to $R_t$ must be forced into knapsack $t$ in non-decreasing order of values ($p_{mk} \times \lambda_{mk}$,) inasmuch as $\sum_{m \in R_t} p_{mk}\lambda_{mk} \le \theta_t$. The improved version of EDM procedure will be obtained by simply substituting (15) instead of (14) in Step (c). To solve (P), the EDM procedure is solved separately for each skill. The integration of the obtained partial schedules yields a near-optimal solution for (P) as will be demonstrated in the next section.

**COMPUTATIONAL EXPERIMENTS**

The applicability and performance of the proposed solution procedure, EDM, is verified by 212 benchmark problems extracted from 35 original datasets associated with the case study described in Section 2. The original datasets are randomly selected from a pool of historical datasets, each of which represents the maintenance and repair jobs of different equipment in various MAs, submitted to the maintenance department during one week. Each benchmark problem is obtained by perturbing an original dataset in terms of the workforce availability. The number of jobs and skills for original datasets ranges between [10, 229] jobs and [7, 34] skills. The maintenance department is responsible for providing the labour resources with the required skills. These may come from internal and external sources and will perform the jobs while minimizing equipment downtime or equivalently minimizing TWCT of the jobs [1]. We use the total number of operations, $OP = \sum_{k=1}^{K} \sum_{m=1}^{M} \text{sgn}(p_{mk})$, as an indicator to measure the size of each dataset. Under the specific case, $OP = M$, each job needs just one skill and thereby (P) is simply decomposed into $K$ single-skill SPSC, as pointed out earlier.

To extract different benchmark problems, various levels of workforce availability are considered. That is, given a dataset consisting of $M$ jobs and requiring $K$ skills, $(M, K)$, the possible minimum and maximum workforce requirements to perform all jobs by all skills are obtained as $W^{\min} = \sum_{k=1}^{K} \max_{1 \leq m \leq M} \{\lambda_{mk}\}$ and $W^{\max} = \sum_{k=1}^{K} \sum_{m=1}^{M} \lambda_{mk}$ respectively. The benchmark scenarios are generated assuming different levels of workforce availability for a given dataset as follows:

$$A_\alpha = \lceil (1-\alpha)W^{\min} + \alpha W^{\max} \rceil,$$

where $\alpha$ = 0.01, 0.05, 0.1, 0,2, 0.3, 0.4, 0.5 . In fact, $A_\alpha$ represents a Sensitivity Analysis (SA) factor so that the smaller the value of $\alpha$, the smaller the workforce size available and therefore, more complicated scenario from a computational perspective due to the resource scarcity. More precisely, considering $\alpha \leq 0.5$, we focus on harder scenarios in order to better evaluate the performance of the EDM.

Through this previously explained pre-processing, we try to keep the diversity of problems high enough in terms of workforce size, number of work orders, processing times and skill requirements. The benchmark problems are optimally solved using the Branch-and-Bound (B&B) method embedded in LINGO 10.0 software on an x64-based DELL workstation with 8 Intel Xenon processors 2.0 GHz and 2 GB memory. The comparison between the B&B and EDM results are summarized in Table 1 in which the optimal objective function value, Z*, is reported if it can be achieved within 3 hours; otherwise, the best and lower bound values obtained in three hours, i.e., $Z^{BEST}$ and $Z^{LB}$, will be reported. The CPU time of EDM is ignored since it is practically negligible compared to the exact method, about 5 seconds in the worst case. In contrary, for medium and large size instances, B&B takes about several minutes to access the feasible space, especially when the labour resource is reduced. The findings show that the percentage gap between $Z^*$ ($Z^{BEST}$) and the objective function value resulted from EDM, $Z^{EDM}$, follows a Weibull distribution with shape parameter $\beta=0.83$ and scale parameter $\eta=0.24$, as graphically shown in Figure 2. This means the average gap between the optimal solution and EDM is about $\eta\Gamma(1+(1/\beta))= 0.26\%$ with a standard deviation of about $\left(\eta^2\left[\Gamma(1+(2/\beta))-\Gamma^2(1+(1/\beta))\right]\right)^{1/2}=0.29$. This is quite a satisfactory outcome, making EDM an efficient method to solve (P). However, in very few cases, 3 out of 212, EDM failed to access the feasible space. That is, the completion time of some jobs exceed the planning horizon length. While such a situation is common in practice, we theoretically consider it as an infeasible situation as long as B&B can produce a schedule bounded by the horizon length. This assumption provides a fair comparison between our methodology and B&B. These three cases can be found as scenario numbers 164, 177, and 212 in Table 1.

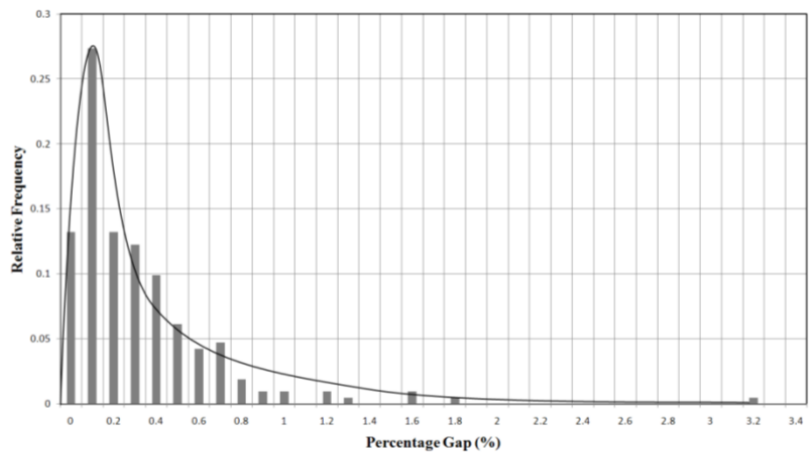

**Figure 2.** Relative frequency of gap between $Z^*$ ($Z^{BEST}$) and $Z^{EDM}$

To illustrate whether or not the size of the scenario affects the quality of the EDM, we arrange the scenarios in a non-decreasing order of ratios $\left\lceil (OP\times K)/A_\alpha^{0.1} \right\rceil$ (horizontal axis - Figure 3). This ad-hoc ratio is used to achieve the maximum number of unique scale indices assigned to the scenarios. The correlation between the scenario scale and the relative gap is calculated to be around +0.21, indicating that there is no significant correlation between problem size and solution quality of the EDM. In other words, increasing the problem size does not significantly affect the relative gap, as evident in Figure 3.

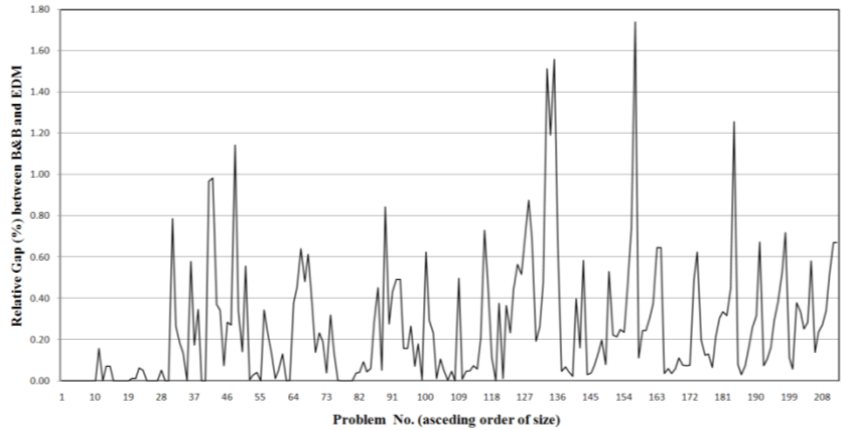

**Figure 3.** Correlation between the scenario size and the relative gap

However, considering each individual dataset, decreasing the workforce size negatively influences both B&B CPU time and the relative gap, e.g., scenarios [110 to 115] or [137 to 143] in Table 1. Based on the empirical evidence, when the workforce size decreases, the computational effort for bound computations in B&B progressively increases. In fact, the smaller the workforce size, the lesser the nodes in the decision tree of B&B are fathomed due to the existence of a weak bind. An important point here is that by reducing the workforce size, the feasible region is tightening intensively. As a consequence, finding a strong bind becomes more difficult especially when the penalization strategy is used to generate the bind. However, this behaviour is not always true, e.g., scenarios [148 to 150] or [195 to 197] in Table 1.

The workforce size reduction also has a negative effect on EDM; however, this is not always the case, e.g., scenarios [173 to 176] in Table 1. The main reason is that skills are considered separately in EDM. Hence, the negative effect of reducing the size of a given skill will not be rectified by other skills. To cope with this drawback, we need to consider a kind of interaction between single-skilled sub-problems, such as in Benders decomposition method. However, any improvement in EDM significantly increases its computational effort that results in its inefficiency as a sensitivity analysis tool. This is not a major concern as we seldom encounter the worst case scenarios of the workforce unavailability in the real world.

Even though the instances are coming from a single plant, they are actually related to completely different equipment types with various maintenance programs and failure modes. Having this, in addition to enough permutations on workforce size generated using $A_\alpha$ factor, the instances are fairly distributed in terms of input parameters. Hence, our statistical analysis on the proposed EDM method can be extended to other similar industries with high confidence.

**Table 1.** Comparison between B&B and EDM

| Problem Info. | | | | | Branch & Bound | | | EDM | |
|---|---|---|---|---|---|---|---|---|---|
| No. | M | OP | K | $A_\alpha$ (SA Factor) | $Z^{LB}$ | $Z^*$ ($Z^{BEST}$) | CPU (sec.) | $Z^{EDM}$ | Gap (%) |
| 1 | 10 | 11 | 7 | 18 | - | 78.4614 | 1 | 78.461403 | 0.00 |
| 2 | 10 | 11 | 7 | 16 | - | 78.5714 | 1 | 78.571404 | 0.00 |
| 3 | 10 | 11 | 7 | 15 | - | 78.5714 | 1 | 78.571404 | 0.00 |
| 4 | 10 | 11 | 7 | 14 | - | 79.3924 | 1 | 79.392403 | 0.00 |
| 5 | 12 | 12 | 6 | 17 | - | 36.8807 | 1 | 36.880699 | 0.00 |
| 6 | 12 | 12 | 6 | 16 | - | 37.0067 | 1 | 37.006699 | 0.00 |
| 7 | 12 | 12 | 6 | 14 | - | 37.0067 | 1 | 37.006699 | 0.00 |
| 8 | 12 | 12 | 6 | 13 | - | 40.0167 | 1 | 40.016701 | 0.00 |
| 9 | 12 | 14 | 8 | 19 | - | 61.9664 | 1 | 61.9664 | 0.00 |
| 10 | 12 | 14 | 8 | 18 | - | 61.9898 | 1 | 61.989799 | 0.00 |
| 11 | 12 | 14 | 8 | 17 | - | 62.0288 | 1 | 62.125198 | 0.16 |
| 12 | 12 | 14 | 8 | 16 | - | 62.2448 | 1 | 62.244801 | 0.00 |
| 13 | 15 | 15 | 7 | 25 | - | 57.8768 | 1 | 57.917198 | 0.07 |
| 14 | 15 | 15 | 7 | 22 | - | 57.9406 | 1 | 57.980999 | 0.07 |
| 15 | 15 | 15 | 7 | 20 | - | 60.176 | 1 | 60.175999 | 0.00 |
| 16 | 15 | 15 | 7 | 16 | - | 60.2163 | 1 | 60.216301 | 0.00 |
| 17 | 14 | 16 | 13 | 30 | - | 48.0565 | 1 | 48.056499 | 0.00 |
| 18 | 14 | 16 | 13 | 29 | - | 48.2815 | 1 | 48.281502 | 0.00 |
| 19 | 14 | 16 | 13 | 28 | - | 48.2815 | 1 | 48.281502 | 0.00 |
| 20 | 13 | 17 | 8 | 25 | - | 79.5361 | 1 | 79.544403 | 0.01 |
| 21 | 13 | 17 | 8 | 24 | - | 79.5859 | 1 | 79.594193 | 0.01 |
| 22 | 13 | 17 | 8 | 22 | - | 79.644 | 1 | 79.693787 | 0.06 |
| 23 | 13 | 17 | 8 | 18 | - | 79.7606 | 1 | 79.802094 | 0.05 |
| 24 | 13 | 17 | 13 | 32 | - | 63.9561 | 1 | 63.956089 | 0.00 |
| 25 | 13 | 17 | 13 | 31 | - | 64.0041 | 1 | 64.0041 | 0.00 |
| 26 | 13 | 17 | 13 | 29 | - | 64.0041 | 1 | 64.0041 | 0.00 |
| 27 | 13 | 17 | 13 | 28 | - | 64.0041 | 1 | 64.0041 | 0.00 |
| 28 | 16 | 18 | 11 | 27 | - | 30.5924 | 1 | 30.6084 | 0.05 |
| 29 | 16 | 18 | 11 | 24 | - | 30.9713 | 1 | 30.9713 | 0.00 |
| 30 | 16 | 18 | 11 | 23 | - | 31.1043 | 1 | 31.1043 | 0.00 |
| 31 | 16 | 18 | 11 | 22 | - | 31.4977 | 1 | 31.7451 | 0.79 |
| 32 | 11 | 21 | 10 | 23 | - | 41.935 | 1 | 42.04591 | 0.26 |
| 33 | 11 | 21 | 10 | 18 | - | 42.0535 | 1 | 42.1299 | 0.18 |
| 34 | 11 | 21 | 10 | 17 | - | 42.1604 | 1 | 42.217701 | 0.14 |
| 35 | 11 | 21 | 10 | 16 | - | 42.4125 | 1 | 42.412498 | 0.00 |
| 36 | 11 | 21 | 10 | 14 | - | 42.8782 | 1 | 43.126202 | 0.58 |
| 37 | 11 | 21 | 10 | 13 | - | 44.3441 | 1 | 44.420502 | 0.17 |
| 38 | 23 | 26 | 8 | 28 | - | 71.4546 | 1 | 71.701103 | 0.34 |
| 39 | 23 | 26 | 8 | 26 | - | 71.8932 | 1 | 71.893204 | 0.00 |
| 40 | 23 | 26 | 8 | 23 | - | 71.9103 | 1 | 71.910301 | 0.00 |
| 41 | 23 | 26 | 8 | 19 | - | 72.10845 | 1 | 72.805 | 0.97 |
| 42 | 23 | 26 | 8 | 16 | - | 72.16155 | 1 | 72.8713 | 0.98 |
| 43 | 23 | 26 | 8 | 15 | - | 72.6108 | 1 | 72.878899 | 0.37 |
| 44 | 23 | 26 | 8 | 14 | - | 72.6925 | 1 | 72.941597 | 0.34 |
| 45 | 23 | 30 | 12 | 38 | - | 48.2528 | 1 | 48.287399 | 0.07 |
| 46 | 23 | 30 | 12 | 33 | - | 49.7256 | 1 | 49.866199 | 0.28 |
| 47 | 23 | 30 | 12 | 31 | - | 49.9312 | 1 | 50.066391 | 0.27 |
| 48 | 23 | 30 | 12 | 27 | - | 51.47185 | 4 | 52.059799 | 1.14 |
| 49 | 23 | 30 | 12 | 22 | - | 52.89388 | 30 | 53.072601 | 0.34 |
| 50 | 23 | 30 | 12 | 21 | - | 54.1766 | 3 | 54.252201 | 0.14 |
| 51 | 23 | 30 | 12 | 20 | - | 55.673 | 2 | 55.982399 | 0.56 |
| 52 | 24 | 30 | 14 | 38 | - | 68.6681 | 1 | 68.671303 | 0.00 |
| 53 | 24 | 30 | 14 | 34 | - | 68.6889 | 1 | 68.708099 | 0.03 |
| 54 | 24 | 30 | 14 | 33 | - | 68.7433 | 1 | 68.772102 | 0.04 |
| 55 | 24 | 30 | 14 | 29 | - | 69.0969 | 1 | 69.096901 | 0.00 |
| 56 | 24 | 30 | 14 | 27 | - | 69.2105 | 1 | 69.447304 | 0.34 |
| 57 | 24 | 30 | 14 | 25 | - | 69.4537 | 1 | 69.6073 | 0.22 |
| 58 | 22 | 31 | 14 | 36 | - | 43.6782 | 1 | 43.738701 | 0.14 |
| 59 | 22 | 31 | 14 | 33 | - | 43.8609 | 1 | 43.866199 | 0.01 |
| 60 | 22 | 31 | 14 | 31 | - | 43.9602 | 1 | 43.982899 | 0.05 |
| 61 | 22 | 31 | 14 | 29 | - | 44.4136 | 6 | 44.4711 | 0.13 |
| 62 | 22 | 31 | 14 | 24 | - | 45.3223 | 13 | 45.3223 | 0.00 |
| 63 | 22 | 31 | 14 | 23 | - | 46.64565 | 1 | 46.646999 | 0.00 |
| 64 | 26 | 33 | 13 | 32 | - | 33.7077 | 1 | 33.835602 | 0.38 |
| 65 | 26 | 33 | 13 | 28 | - | 33.8009 | 1 | 33.954102 | 0.45 |
| 66 | 26 | 33 | 13 | 26 | - | 33.8939 | 1 | 34.111198 | 0.64 |
| 67 | 26 | 33 | 13 | 25 | - | 34.0034 | 1 | 34.167099 | 0.48 |
| 68 | 26 | 33 | 13 | 21 | - | 34.5891 | 5 | 34.801201 | 0.61 |
| 69 | 26 | 33 | 13 | 20 | - | 35.2433 | 74 | 35.373299 | 0.37 |
| 70 | 25 | 34 | 11 | 35 | - | 64.1962 | 1 | 64.285301 | 0.14 |
| 71 | 25 | 34 | 11 | 31 | - | 64.2804 | 1 | 64.429314 | 0.23 |
| 72 | 25 | 34 | 11 | 28 | - | 64.7903 | 1 | 64.914787 | 0.19 |
| 73 | 25 | 34 | 11 | 26 | - | 65.7066 | 6 | 65.731194 | 0.04 |
| 74 | 25 | 34 | 11 | 24 | - | 66.7924 | 1 | 67.005997 | 0.32 |
| 75 | 25 | 34 | 11 | 20 | - | 67.6161 | 5 | 67.713997 | 0.14 |
| 76 | 26 | 34 | 11 | 38 | - | 51.3759 | 1 | 51.376801 | 0.00 |
| 77 | 26 | 34 | 11 | 35 | - | 52.427 | 26 | 52.426998 | 0.00 |
| 78 | 26 | 34 | 11 | 32 | - | 52.4861 | 1 | 52.486099 | 0.00 |
| 79 | 26 | 34 | 11 | 27 | - | 54.8165 | 1 | 54.817101 | 0.00 |
| 80  | 26  | 34  | 11 | 24  | -       | 56.685   | 1      | 58.4436   | 3.10 |
| 81  | 26  | 34  | 11 | 22  | -       | 58.472   | 1      | 58.4949   | 0.04 |
| 82  | 26  | 34  | 11 | 21  | -       | 64.8353  | 1      | 70.191513 | 8.26 |
| 83  | 31  | 34  | 8  | 38  | -       | 10.4042  | 1      | 10.4139   | 0.09 |
| 84  | 31  | 34  | 8  | 34  | -       | 10.4362  | 1      | 10.4407   | 0.04 |
| 85  | 31  | 34  | 8  | 30  | -       | 10.4614  | 1      | 10.4677   | 0.06 |
| 86  | 31  | 34  | 8  | 24  | -       | 10.5136  | 1      | 10.5431   | 0.28 |
| 87  | 31  | 34  | 8  | 20  | -       | 10.6225  | 83     | 10.6705   | 0.45 |
| 88  | 31  | 34  | 8  | 18  | -       | 10.77937 | 2      | 10.785    | 0.05 |
| 89  | 31  | 34  | 8  | 16  | -       | 11.069   | 19     | 11.1622   | 0.84 |
| 90  | 25  | 36  | 14 | 38  | -       | 31.0453  | 1      | 31.131201 | 0.28 |
| 91  | 25  | 36  | 14 | 32  | -       | 32.1786  | 1      | 32.318001 | 0.43 |
| 92  | 25  | 36  | 14 | 30  | -       | 32.6185  | 1      | 32.779301 | 0.49 |
| 93  | 25  | 36  | 14 | 26  | -       | 32.724   | 1      | 32.8848   | 0.49 |
| 94  | 25  | 36  | 14 | 22  | -       | 34.2363  | 1      | 34.289799 | 0.16 |
| 95  | 25  | 36  | 14 | 19  | -       | 34.2898  | 1      | 34.3433   | 0.16 |
| 96  | 33  | 37  | 13 | 45  | -       | 44.9073  | 1      | 45.0257   | 0.26 |
| 97  | 33  | 37  | 13 | 42  | -       | 44.9935  | 1      | 45.0257   | 0.07 |
| 98  | 33  | 37  | 13 | 39  | -       | 45.2286  | 2      | 45.3088   | 0.18 |
| 99  | 33  | 37  | 13 | 33  | -       | 45.6858  | 2      | 45.688    | 0.00 |
| 100 | 33  | 37  | 13 | 27  | -       | 47.041   | 3      | 47.335011 | 0.63 |
| 101 | 33  | 37  | 13 | 26  | -       | 47.6374  | 10     | 47.7766   | 0.29 |
| 102 | 33  | 37  | 13 | 25  | -       | 48.0298  | 23     | 48.141411 | 0.23 |
| 103 | 34  | 38  | 10 | 38  | -       | 62.63855 | 1      | 62.644508 | 0.01 |
| 104 | 34  | 38  | 10 | 34  | -       | 62.71377 | 1      | 62.779709 | 0.11 |
| 105 | 34  | 38  | 10 | 30  | -       | 62.9434  | 1      | 62.976009 | 0.05 |
| 106 | 34  | 38  | 10 | 25  | -       | 63.4971  | 1      | 63.4981   | 0.00 |
| 107 | 34  | 38  | 10 | 21  | -       | 64.1119  | 1      | 64.140511 | 0.04 |
| 108 | 34  | 38  | 10 | 19  | -       | 64.6039  | 2      | 64.603897 | 0.00 |
| 109 | 34  | 38  | 10 | 17  | -       | 65.5329  | 2      | 65.859123 | 0.50 |
| 110 | 33  | 39  | 17 | 46  | -       | 63.6819  | 1      | 63.686897 | 0.01 |
| 111 | 33  | 39  | 17 | 41  | -       | 65.3223  | 1      | 65.352303 | 0.05 |
| 112 | 33  | 39  | 17 | 39  | -       | 65.3328  | 1      | 65.365311 | 0.05 |
| 113 | 33  | 39  | 17 | 36  | -       | 65.371   | 1      | 65.419006 | 0.07 |
| 114 | 33  | 39  | 17 | 34  | -       | 65.41    | 1      | 65.448013 | 0.06 |
| 115 | 33  | 39  | 17 | 32  | -       | 65.5285  | 2      | 65.6605   | 0.20 |
| 116 | 37  | 43  | 15 | 60  | -       | 49.50492 | 1      | 49.865799 | 0.73 |
| 117 | 37  | 43  | 15 | 57  | -       | 49.7467  | 1      | 49.978588 | 0.47 |
| 118 | 37  | 43  | 15 | 53  | -       | 50.4195  | 7      | 50.47739  | 0.11 |
| 119 | 37  | 43  | 15 | 45  | -       | 51.6892  | 2      | 51.689201 | 0.00 |
| 120 | 37  | 43  | 15 | 41  | -       | 52.87972 | 2      | 53.077789 | 0.37 |
| 121 | 37  | 43  | 15 | 38  | -       | 53.9686  | 8      | 53.97459  | 0.01 |
| 122 | 37  | 43  | 15 | 37  | -       | 54.28123 | 107    | 54.478588 | 0.36 |
| 123 | 28  | 48  | 13 | 45  | -       | 38.8189  | 1      | 38.909302 | 0.23 |
| 124 | 28  | 48  | 13 | 39  | -       | 38.8839  | 1      | 39.056801 | 0.44 |
| 125 | 28  | 48  | 13 | 36  | -       | 38.9804  | 7      | 39.200401 | 0.56 |
| 126 | 28  | 48  | 13 | 30  | -       | 39.2481  | 142    | 39.450199 | 0.51 |
| 127 | 28  | 48  | 13 | 27  | -       | 39.6097  | 140    | 39.880699 | 0.68 |
| 128 | 28  | 48  | 13 | 26  | -       | 39.6905  | 644    | 40.0378   | 0.88 |
| 129 | 28  | 48  | 13 | 23  | -       | 40.6617  | 6176   | 40.940399 | 0.69 |
| 130 | 57  | 63  | 18 | 75  | -       | 51.4336  | 1      | 51.532799 | 0.19 |
| 131 | 57  | 63  | 18 | 66  | -       | 52.08743 | 1      | 52.223709 | 0.26 |
| 132 | 57  | 63  | 18 | 61  | -       | 52.51772 | 1      | 52.769211 | 0.48 |
| 133 | 57  | 63  | 18 | 52  | -       | 53.91232 | 1      | 54.728512 | 1.51 |
| 134 | 57  | 63  | 18 | 44  | -       | 55.7953  | 990    | 56.460209 | 1.19 |
| 135 | 57  | 63  | 18 | 42  | -       | 56.8174  | 681    | 57.70369  | 1.56 |
| 136 | 57  | 63  | 18 | 39  | -       | 59.723   | 8140   | 60.132309 | 0.69 |
| 137 | 61  | 71  | 15 | 61  | -       | 61.2602  | 2      | 61.288109 | 0.05 |
| 138 | 61  | 71  | 15 | 54  | -       | 61.3916  | 45     | 61.432812 | 0.07 |
| 139 | 61  | 71  | 15 | 46  | -       | 61.5675  | 3      | 61.5937   | 0.04 |
| 140 | 61  | 71  | 15 | 35  | -       | 61.8864  | 6      | 61.89909  | 0.02 |
| 141 | 61  | 71  | 15 | 28  | -       | 62.7714  | 123    | 63.021389 | 0.40 |
| 142 | 61  | 71  | 15 | 24  | -       | 63.9092  | 173    | 64.010689 | 0.16 |
| 143 | 61  | 71  | 15 | 21  | -       | 65.0649  | 191    | 65.444878 | 0.58 |
| 144 | 79  | 88  | 20 | 89  | -       | 100.3176 | 4      | 100.34721 | 0.03 |
| 145 | 79  | 88  | 20 | 76  | -       | 100.5283 | 8      | 100.56611 | 0.04 |
| 146 | 79  | 88  | 20 | 65  | -       | 100.7591 | 3      | 100.83481 | 0.08 |
| 147 | 79  | 88  | 20 | 52  | -       | 101.2413 | 11     | 101.38219 | 0.14 |
| 148 | 79  | 88  | 20 | 43  | -       | 102.3672 | 1032   | 102.56953 | 0.20 |
| 149 | 79  | 88  | 20 | 35  | -       | 103.9585 | 44     | 104.04023 | 0.08 |
| 150 | 79  | 88  | 20 | 31  | -       | 107.8838 | 15579  | 108.45473 | 0.53 |
| 151 | 101 | 105 | 17 | 114 | -       | 58.4826  | 4      | 58.612591 | 0.22 |
| 152 | 101 | 105 | 17 | 100 | -       | 58.6375  | 6      | 58.762081 | 0.21 |
| 153 | 101 | 105 | 17 | 85  | -       | 58.9591  | 3      | 59.10508  | 0.25 |
| 154 | 101 | 105 | 17 | 69  | -       | 59.5387  | 43     | 59.67799  | 0.23 |
| 155 | 101 | 105 | 17 | 55  | -       | 60.6508  | 241    | 60.92329  | 0.45 |
| 156 | 101 | 105 | 17 | 46  | -       | 62.02551 | 699    | 62.484581 | 0.74 |
| 157 | 101 | 105 | 17 | 40  | -       | 64.2455  | 467    | 65.363487 | 1.74 |
| 158 | 135 | 144 | 21 | 122 | -       | 34.9864  | 5      | 35.025101 | 0.11 |
| 159 | 135 | 144 | 21 | 107 | -       | 35.1233  | 13     | 35.208511 | 0.24 |
| 160 | 135 | 144 | 21 | 89  | -       | 35.4633  | 22     | 35.54921  | 0.24 |
| 161 | 135 | 144 | 21 | 73  | -       | 36.1151  | 50     | 36.2258   | 0.31 |
| 162 | 135 | 144 | 21 | 55  | -       | 37.7061  | 140    | 37.847488 | 0.37 |
| 163 | 135 | 144 | 21 | 45  | 39.54251| 39.565   | > 3 hrs| 39.820202 | 0.65 |
| 164 | 135 | 144 | 21 | 38 | 45.84522 | 45.8867 | > 3 hrs | N/A | - |
| 165 | 132 | 145 | 25 | 137 | - | 58.6599 | 7 | 58.67989 | 0.03 |
| 166 | 132 | 145 | 25 | 114 | - | 58.8208 | 9 | 58.856499 | 0.06 |
| 167 | 132 | 145 | 25 | 97 | - | 58.6599 | 8 | 58.67989 | 0.03 |
| 168 | 132 | 145 | 25 | 79 | - | 58.8208 | 9 | 58.856499 | 0.06 |
| 169 | 132 | 145 | 25 | 63 | - | 59.1084 | 37 | 59.1745 | 0.11 |
| 170 | 132 | 145 | 25 | 46 | - | 61.0297 | 68 | 61.075489 | 0.08 |
| 171 | 141 | 159 | 29 | 147 | - | 61.47041 | 6 | 61.51598 | 0.07 |
| 172 | 141 | 159 | 29 | 126 | - | 61.5696 | 119 | 61.615971 | 0.08 |
| 173 | 141 | 159 | 29 | 106 | - | 61.88495 | 12 | 62.187 | 0.49 |
| 174 | 141 | 159 | 29 | 87 | - | 62.37121 | 15 | 62.76041 | 0.62 |
| 175 | 141 | 159 | 29 | 68 | - | 63.6624 | 701 | 63.788502 | 0.20 |
| 176 | 141 | 159 | 29 | 56 | - | 65.1911 | > 3 hrs | 65.272812 | 0.13 |
| 177 | 141 | 159 | 29 | 49 | 67.47359 | 67.4839 | > 3 hrs | N/A | - |
| 178 | 145 | 160 | 20 | 147 | - | 48.6565 | 11 | 48.68819 | 0.07 |
| 179 | 145 | 160 | 20 | 123 | - | 48.8381 | 52 | 48.941078 | 0.21 |
| 180 | 145 | 160 | 20 | 102 | - | 49.1908 | 437 | 49.34079 | 0.30 |
| 181 | 145 | 160 | 20 | 82 | - | 49.8687 | 9863 | 50.035702 | 0.33 |
| 182 | 145 | 160 | 20 | 60 | - | 51.4081 | 25371 | 51.570789 | 0.32 |
| 183 | 145 | 160 | 20 | 50 | 53.01927 | 53.0252 | > 3 hrs | 53.262699 | 0.45 |
| 184 | 145 | 160 | 20 | 39 | 57.95624 | 58.0672 | > 3 hrs | 58.797199 | 1.26 |
| 185 | 152 | 163 | 23 | 156 | - | 68.8081 | 13 | 68.861481 | 0.08 |
| 186 | 152 | 163 | 23 | 134 | - | 68.9753 | 156 | 68.995491 | 0.03 |
| 187 | 152 | 163 | 23 | 114 | - | 69.2349 | 2942 | 69.284691 | 0.07 |
| 188 | 152 | 163 | 23 | 92 | - | 69.7629 | 3480 | 69.87291 | 0.16 |
| 189 | 152 | 163 | 23 | 68 | - | 71.0437 | 4933 | 71.228104 | 0.26 |
| 190 | 152 | 163 | 23 | 58 | - | 72.3003 | 1573 | 72.528023 | 0.31 |
| 191 | 152 | 163 | 23 | 47 | 75.53397 | 75.5825 | > 3 hrs | 76.090736 | 0.67 |
| 192 | 169 | 181 | 22 | 164 | - | 67.5518 | 11 | 67.601112 | 0.07 |
| 193 | 169 | 181 | 22 | 139 | - | 67.728 | 299 | 67.797791 | 0.10 |
| 194 | 169 | 181 | 22 | 116 | - | 67.9832 | 1570 | 68.093102 | 0.16 |
| 195 | 169 | 181 | 22 | 93 | - | 68.4199 | 2662 | 68.62178 | 0.30 |
| 196 | 169 | 181 | 22 | 68 | 69.77254 | 69.7765 | > 3 hrs | 70.04612 | 0.39 |
| 197 | 169 | 181 | 22 | 56 | - | 71.2558 | 62792 | 71.628502 | 0.52 |
| 198 | 169 | 181 | 22 | 46 | 75.98684 | 76.0041 | > 3 hrs | 76.549423 | 0.72 |
| 199 | 201 | 217 | 28 | 205 | - | 65.8969 | 136 | 65.969887 | 0.11 |
| 200 | 201 | 217 | 28 | 174 | - | 66.0796 | 207 | 66.117493 | 0.06 |
| 201 | 201 | 217 | 28 | 146 | - | 66.38143 | 12 | 66.632683 | 0.38 |
| 202 | 201 | 217 | 28 | 115 | - | 67.30126 | 20 | 67.526192 | 0.33 |
| 203 | 201 | 217 | 28 | 89 | - | 68.5811 | 23619 | 68.753708 | 0.25 |
| 204 | 201 | 217 | 28 | 74 | 70.20005 | 70.2061 | > 3 hrs | 70.405617 | 0.28 |
| 205 | 201 | 217 | 28 | 61 | - | 73.8265 | 11130 | 74.255943 | 0.58 |
| 206 | 229 | 248 | 34 | 254 | - | 52.6115 | 73 | 52.68362 | 0.14 |
| 207 | 229 | 248 | 34 | 215 | - | 52.8597 | 21 | 52.984501 | 0.24 |
| 208 | 229 | 248 | 34 | 179 | 53.33973 | 53.3436 | > 3 hrs | 53.487801 | 0.27 |
| 209 | 229 | 248 | 34 | 137 | 54.26741 | 54.2793 | > 3 hrs | 54.463711 | 0.34 |
| 210 | 229 | 248 | 34 | 104 | 56.16489 | 56.16781 | > 3 hrs | 56.45932 | 0.52 |
| 211 | 229 | 248 | 34 | 83 | 58.86389 | 58.8848 | > 3 hrs | 59.278702 | 0.67 |
| 212 | 229 | 248 | 34 | 66 | 63.9224 | 64.0133 | > 3 hrs | N/A | - |

**CONCLUSION**

In this paper, we present a quick and efficient solution approach to solving the scheduling problem under the skilled-workforce constraint. In this problem, a set of jobs must be performed by a set of skills, under the workforce size constraint. The proposed approach is an efficient post-optimization tool to conduct the sensitivity analysis as a critical requirement in many industries in which the availability of equipment and maintenance labour resource is a major concern in the presence of unexpected events.. A time-indexed 0-1 mathematical formulation is proposed along with a heuristic solution approach to solve the model. The idea behind the heuristic approach comes from the well-known Dantzig method proposed to solve the Knapsack problem (KP). At first, the proposed model is decomposed into a number of single-skill sub-problems where each one can be presented as a series of nested Binary KP. Then, the Dantzig method is extended to solve the sub-problems. A solution to the primary model is obtained by the integration of the partial schedules associated with these sub-problems. The findings reveal that the proposed approach is able to generate high-quality solutions very close to the optimal ones, in a short amount of time, regardless of the problem size.

The performance of the approach is validated by a number of real datasets from a steel production company. The proposed approach can be used as a quick and efficient tool to conduct the sensitivity analysis in terms of the workforce size as a necessity in maintenance department of many industries. Our heuristic can be developed as an add-in module for assisting maintenance supervisors to quickly re-assign the workforce in the presence of unexpected work orders. Even though there is no theoretical evidence to show that the heuristic method based on the skill decomposition is an $\varepsilon$-approximation algorithm to the considered problem, we statistically show that our proposed method is a $\varepsilon$-approximation algorithm.

Further developments of our approach will address the possibility of considering other factors such as job pre-emption and inter-skill precedence relations. Job pre-emption most likely reduces the complexity of the problem since we no longer need to consider the nested strategy while each period of our planning horizon can be considered as an individual knapsack. The proposed model may be also extended within the framework of *Robust Optimization* so that the tasks' processing times represent a disturbance quantity varying within a structured uncertainty set with an unknown pattern.